\documentclass[final]{cvpr}

\usepackage{times}
\usepackage{epsfig}
\usepackage{graphicx}
\graphicspath{{images/}}
\usepackage{amsmath}
\usepackage{amssymb}

\usepackage{comment}
\usepackage{color}
\usepackage{wrapfig}
\usepackage{tabularx}
\newcolumntype{C}{>{\centering\arraybackslash}X}
\usepackage{stfloats} % make double column table to go bottom
\usepackage[table]{xcolor}
\usepackage{url}
\usepackage{multicol} % making composed figures as tables

\usepackage[pagebackref=true,breaklinks=true,colorlinks,bookmarks=false]{hyperref}

\def\cvprPaperID{19} % *** Enter the CVPR Paper ID here
\def\confYear{CVPR 2021}

\begin{document}

\title{Filter Distribution Templates in Convolutional Networks \\ for Image Classification Tasks}

\author{Ramon Izquierdo-Cordova, Walterio Mayol-Cuevas\\
Department of Computer Science, University of Bristol, United Kingdom\\
{\tt\small \{ramon.izquierdocordova,walterio.mayol-cuevas\}@bristol.ac.uk}
}

\def\AcknowledgmentsText{This work was partially supported by CONACYT and the Secretar\'ia de Educaci\'on P\'ublica, M\'exico.}

%******************
\maketitle

%%%%%%%%%%%%%%%%%%%%%%%%%%%%%%%%%%%%%%%%%%%%%%%%%%%%%%%%%%%%%%%%%%%%%%%%%%%%%%%%
\begin{abstract}

Neural network designers have reached progressive accuracy by increasing models depth, introducing new layer types and discovering new combinations of layers. A common element in many architectures is the distribution of the number of filters in each layer. Neural network models keep a pattern design of increasing filters in deeper layers such as those in LeNet, VGG, ResNet, MobileNet and even in automatic discovered architectures such as NASNet. It remains unknown if this pyramidal distribution of filters is the best for different tasks and constrains. In this work we present a series of modifications in the distribution of filters in four popular neural network models and their effects in accuracy and resource consumption. Results show that by applying this approach, some models improve up to 8.9\% in accuracy showing reductions in parameters up to 54\%.

\end{abstract}

%%%%%%%%%%%%%%%%%%%%%%%%%%%%%%%%%%%%%%%%%%%%%%%%%%%%%%%%%%%%%%%%%%%%%%%%%%%%%%%%
\section{Introduction}

An important consideration to create a convolutional neural network (CNN) model is the number of filters required at every layer. The Neocognitron implementation for example, keeps an equal number of filters for each layer in the model \cite{fukushima1980neocognitron}. A very common practice has been to use a bipyramidal architecture. The number of filters across the different layers is usually increased as the size of the feature maps decreases. This pattern was first proposed in \cite{lecun1998gradient} with the introduction of LeNet and can be observed in a diverse set of models such as VGG\cite{simonyan2014very}, ResNet\cite{he2016deep} and MobileNet\cite{howard2017mobilenets}. Even models obtained from automatic model discovery, like NASNet \cite{zoph2017learning}, follow this principle since neural architecture search methods are mainly formulated to search for layers and connections while the number of filters in each layer remains fixed. The motivation behind this progressive increase in the number of kernels is to compensate a possible loss of the representation caused by the spatial resolution reduction \cite{lecun1998gradient}. In practice it improves performance by keeping a constant number of operations in each layer \cite{chu2014analysis}. It remains unknown if this pyramidal distribution of filters is also beneficial to different aspects of model performances other than the number of operations.

\begin{figure*}
  \begin{multicols}{5}
    \includegraphics[width=0.9\linewidth]{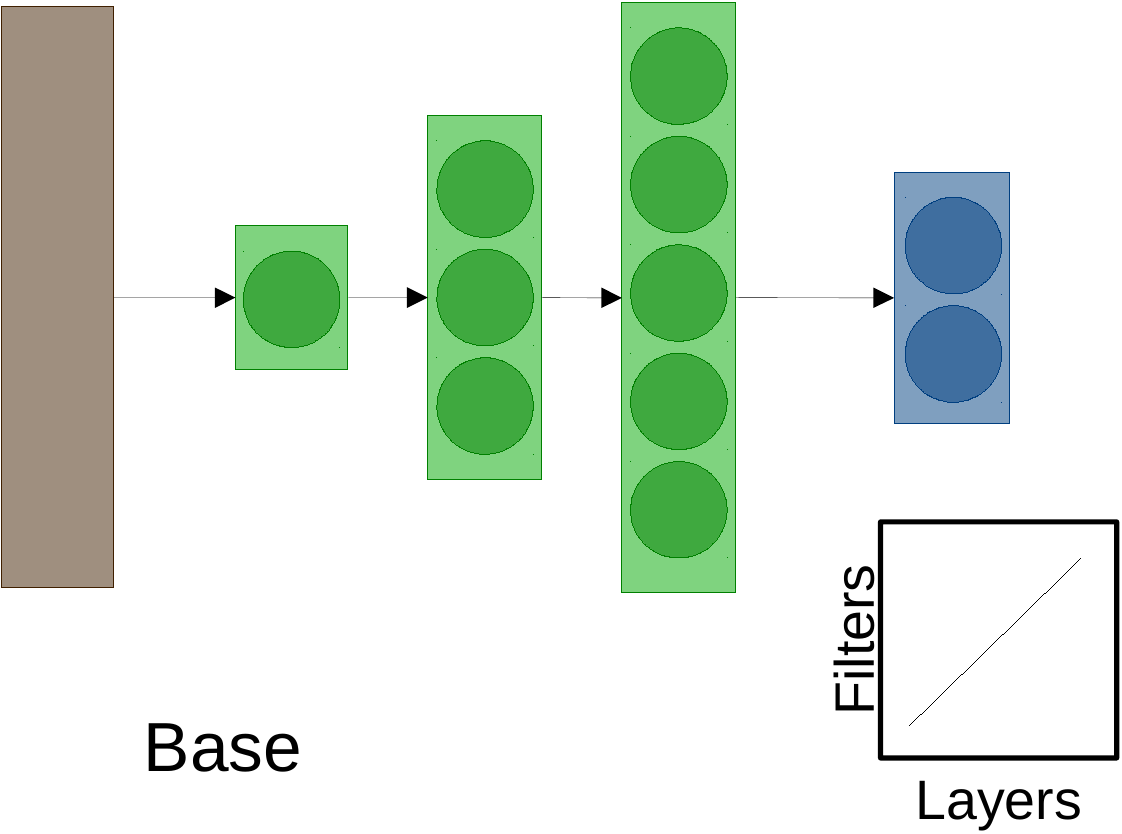}\par
    \includegraphics[width=0.9\linewidth]{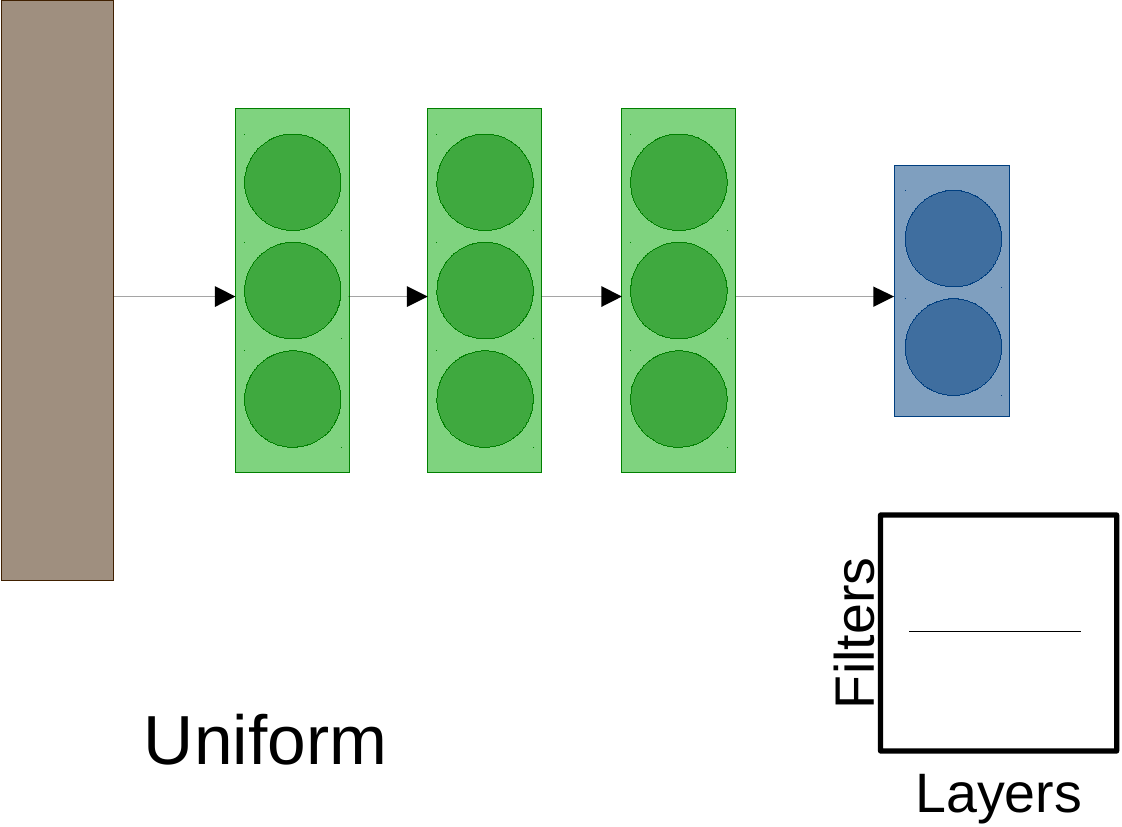}\par
    \includegraphics[width=0.9\linewidth]{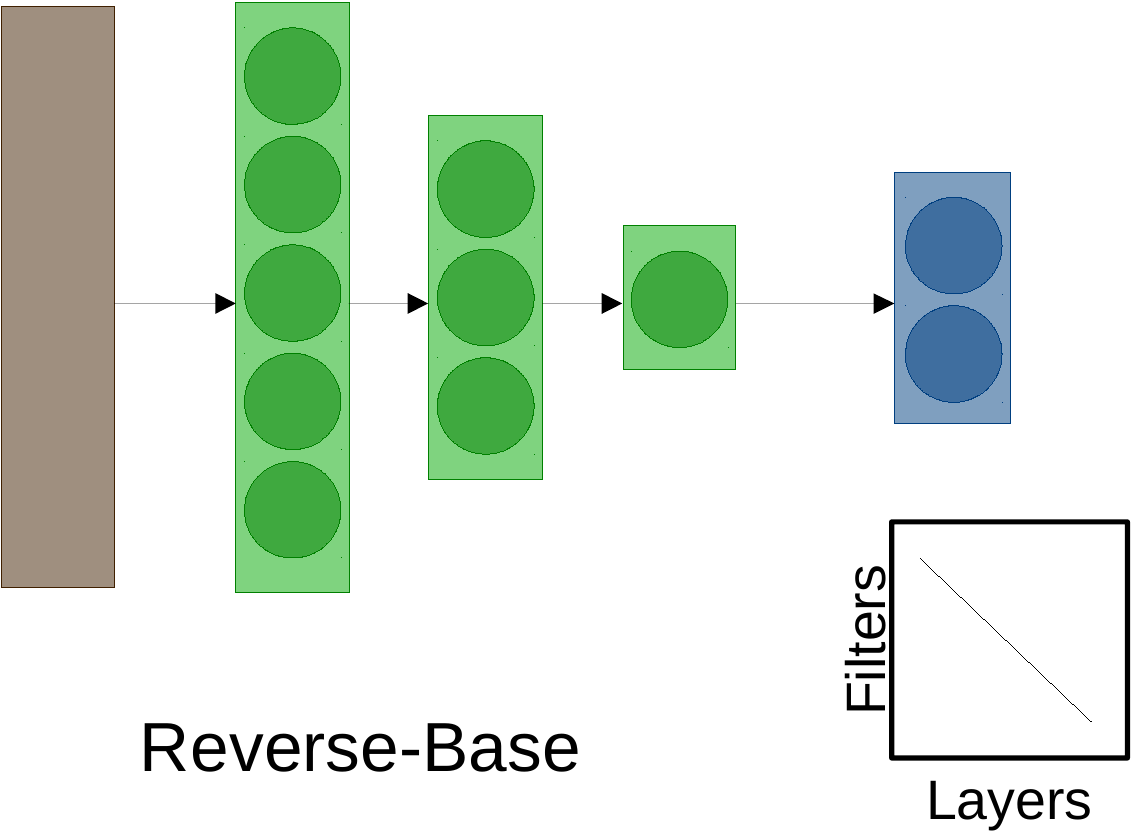}\par
    \includegraphics[width=0.9\linewidth]{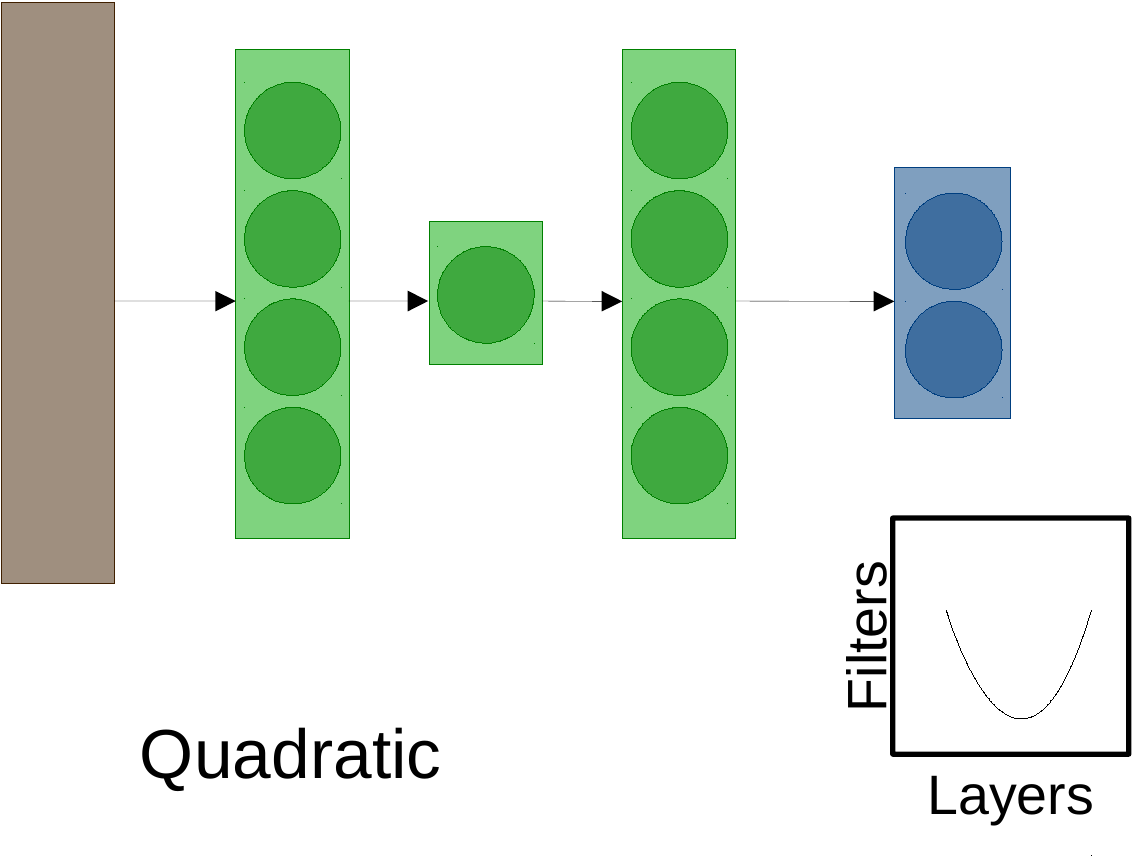}\par
    \includegraphics[width=0.9\linewidth]{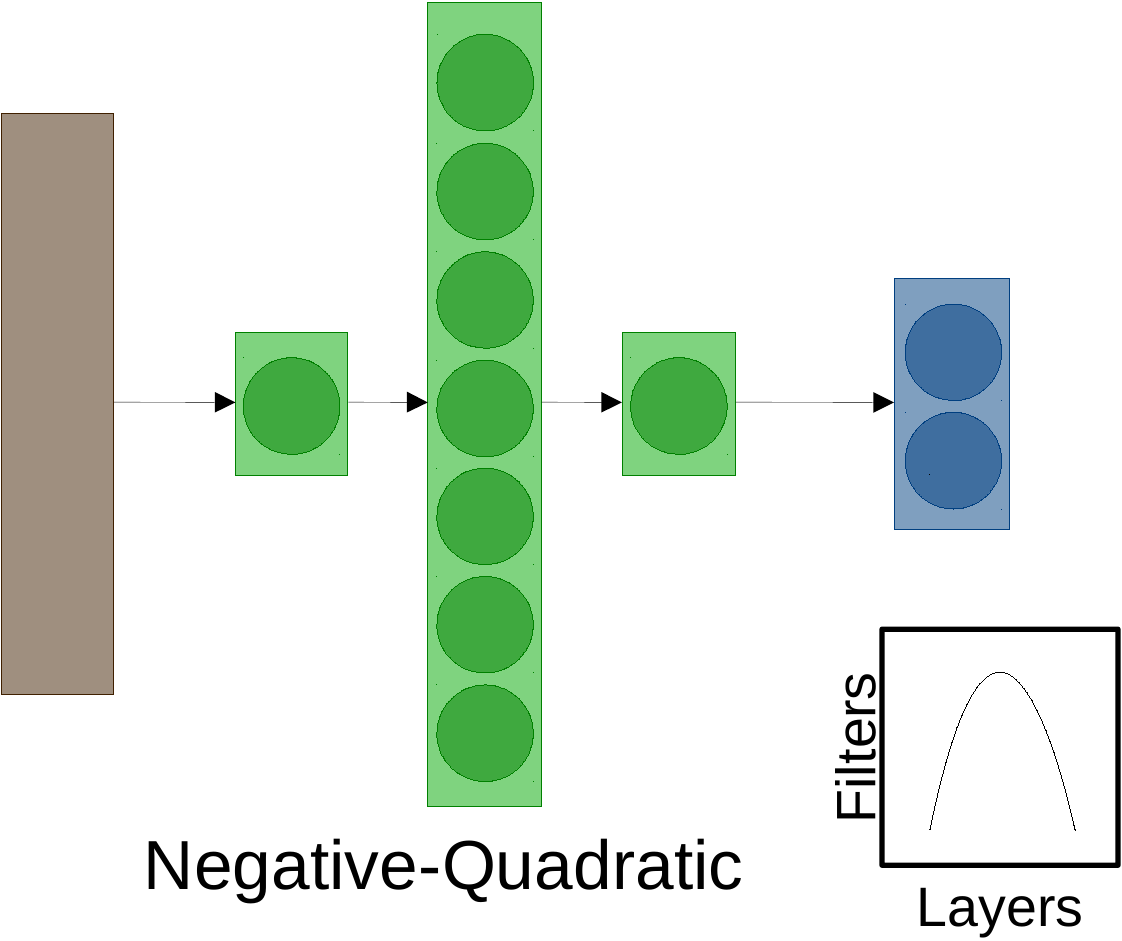}\par
  \end{multicols}
    \caption{Filters per layer using the proposed templates for filter redistribution in a VGG style model. Base distribution, which is the original distribution, shows the common design of growing the filters when resolution of feature maps decreases in deeper layers. Although the total number of filters is kept constant after templates, changes in filter distribution induce different effects in performance and resource consumption.}
\label{fig:templates}
\end{figure*}

The contribution of this paper is to challenge the widely used design of increasing filters in neural convolutional models by applying a small subset of diverse filter distributions, called templates, to existing neural network designs. Experimental evidence shows that simple changes to the pyramidal distribution of filters in convolutional network models lead to improvements in accuracy, number of parameters or memory footprint; we highlight that most recent models, which have had a more detailed tuning in the filter distribution, present resiliency in accuracy to changes in the filter distribution, a phenomena that requires further research and explanation.

Experiments in this document are exploratory. We use equal number of filters in all templates without constraining the effects of the redistribution. We extend this work in \cite{izquierdo2021filter} where templates are evaluated with more rigorous experiments keeping FLOPs to similar values as in the original model and then comparing resource consumption.

%%%%%%%%%%%%%%%%%%%%%%%%%%%%%%%%%%%%%%%%%%%%%%%%%%%%%%%%%%%%%%%%%%%%%%%%%%%%%%%%
\section{Related Work}\label{chap:related}

The process of designing a neural network is a task that has largely been based on experience and experimentation which consumes a lot of time and computational resources. Of note are reference models such as VGG\cite{simonyan2014very}, ResNet\cite{he2016deep}, Inception\cite{szegedy2015going} and MobileNet\cite{howard2017mobilenets} that have been developed with significant use of heuristics. Even with automatic methods, one key feature that has constantly been adopted is the manual selection of the number of filters in each layer in the final model. Filters are set in such a way to have an increasing number as the layers go deeper, differing from the original Neocognitron design\cite{fukushima1980neocognitron}.

With the increase in the use of Neural Networks, and particularly Convolutional Networks for computer vision problems, a mechanism to automatically find the best architecture has become a requirement in the field of Deep Learning. One of the biggest challenges in automatic architecture design is that the search space for CNN architectures is very large\cite{ren2020comprehensive}. Two fields have derived from the problem: \textit{i)} neural architecture search (NAS), that develops mechanisms for searching for the best combination of layers\cite{zoph2018learning} and \textit{ii)} channel number search (CNS), which look for the best distribution of filters given an initial architecture\cite{dong2019network,wang2020revisiting}.

Pruning methods could be seen as an special case of CNS in which there is the assumption that the weights, obtained at the end of the training process of the original model, are important to the pruning method\cite{frankle2018lottery}.

In pruning methods, searching involves training models for several iterations to select the correct weights to remove \cite{frankle2018lottery,he2019filter,you2019gate}, or at least increasing the computation during the training when doing jointly training and search \cite{leclerc2018smallify,li2019learning}. In \cite{liu2018rethinking} it is suggested that accuracy obtained by pruning techniques can be reached by removing filters to fit a certain resource budget and training from scratch.

Our work for finding an appropriate distribution of filters relates to \cite{gordon2018morphnet} in the sense that their method is not restricted to reducing filters but also to increase them to see if the changes are beneficial. Our approach differs however, because it only requires the model to be trained in the final stage, after manually making some predefined changes to the number of filters using the redistribution templates.

%%%%%%%%%%%%%%%%%%%%%%%%%%%%%%%%%%%%%%%%%%%%%%%%%%%%%%%%%%%%%%%%%%%%%%%%%%%%%%%%
\section{Filter distribution templates}\label{chap:methods}

While most of neural network architectures show an incremental distribution of filters, recent pruning methods such as \cite{gordon2018morphnet,leclerc2018smallify}, have shown different filter distribution patterns emerging when reducing models like VGG that defy the notion of pyramidal design as the best distribution for a model. This is a motivational insight into what other distributions can and should be considered when designing models. On one side the combinatorial space of distributions make this a challenging exploration, on the other however, it importantly highlights the need to pursue such exploration if gains in accuracy and overall performance can be made.

In this work, rather than attempting to find the optimal filter distribution with expensive automatic pruning or growing techniques, we propose to first adjust the filters of a convolutional network model via a small number of pre-defined templates. These templates such as those depicted in figure~\ref{fig:templates}, are inspired by existing models that have already been found to perform well and thus candidates that could be beneficial for model performance improvement beyond the number of operations. Performance criteria such as accuracy, memory footprint and inference time are arguably as important as the number of operations required.

In particular, we adopt as one template, a distribution with a fixed number of filters as with the original Neocognitron design, but also other templates inspired by the patterns found in \cite{gordon2018morphnet} where some behaviours are present in different blocks from the resulting ResNet101 model: 1) filters agglomerate in the centre and 2) filters are reduced in the centre of the block. In \cite{leclerc2018smallify,you2019gate} is shown also a pattern with more filters in the centre of a VGG model. Based on these observations we define the templates we use in this work.

\begin{figure}
\centering
  \begin{multicols}{2}
    \includegraphics[width=1.0\linewidth]{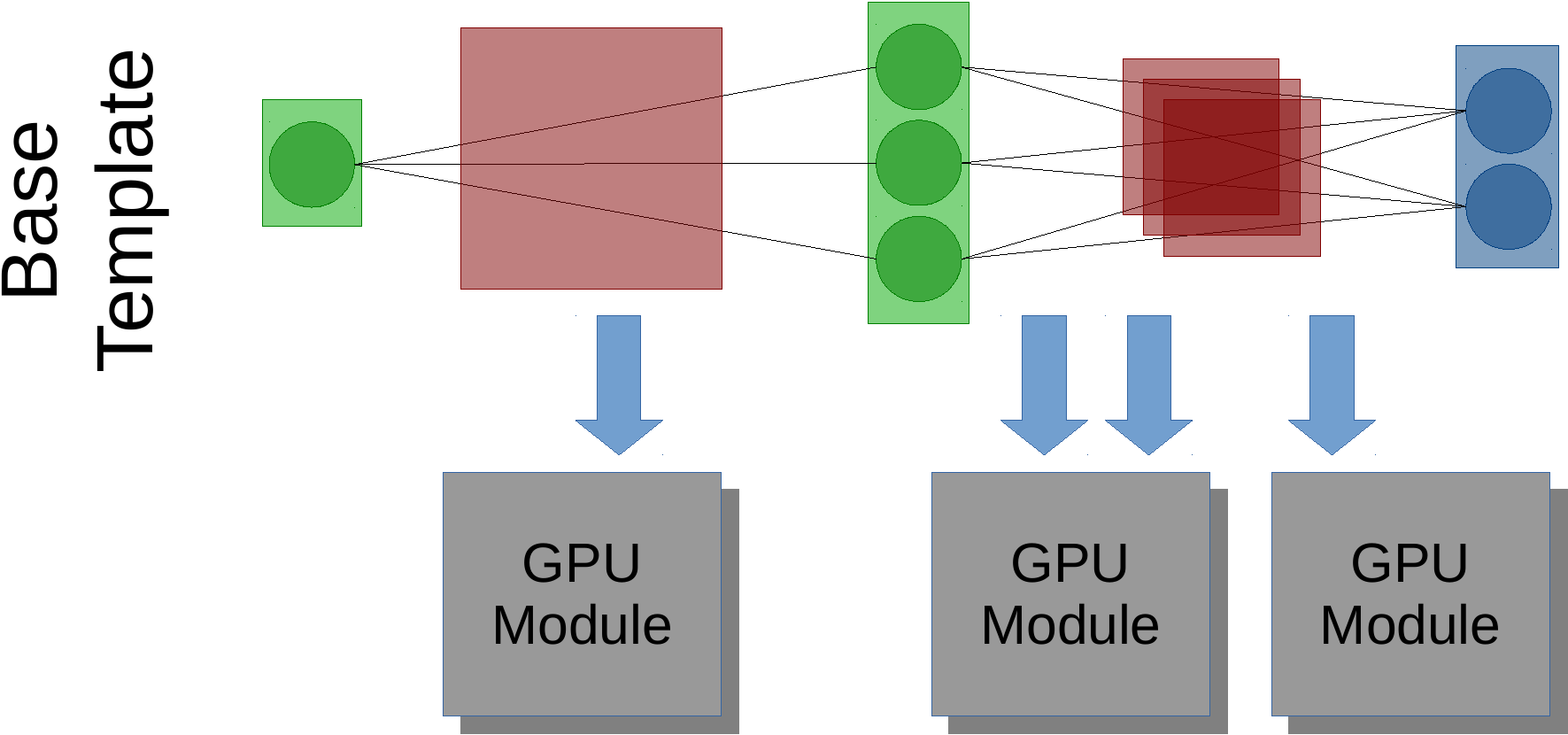}\par
    \includegraphics[width=1.0\linewidth]{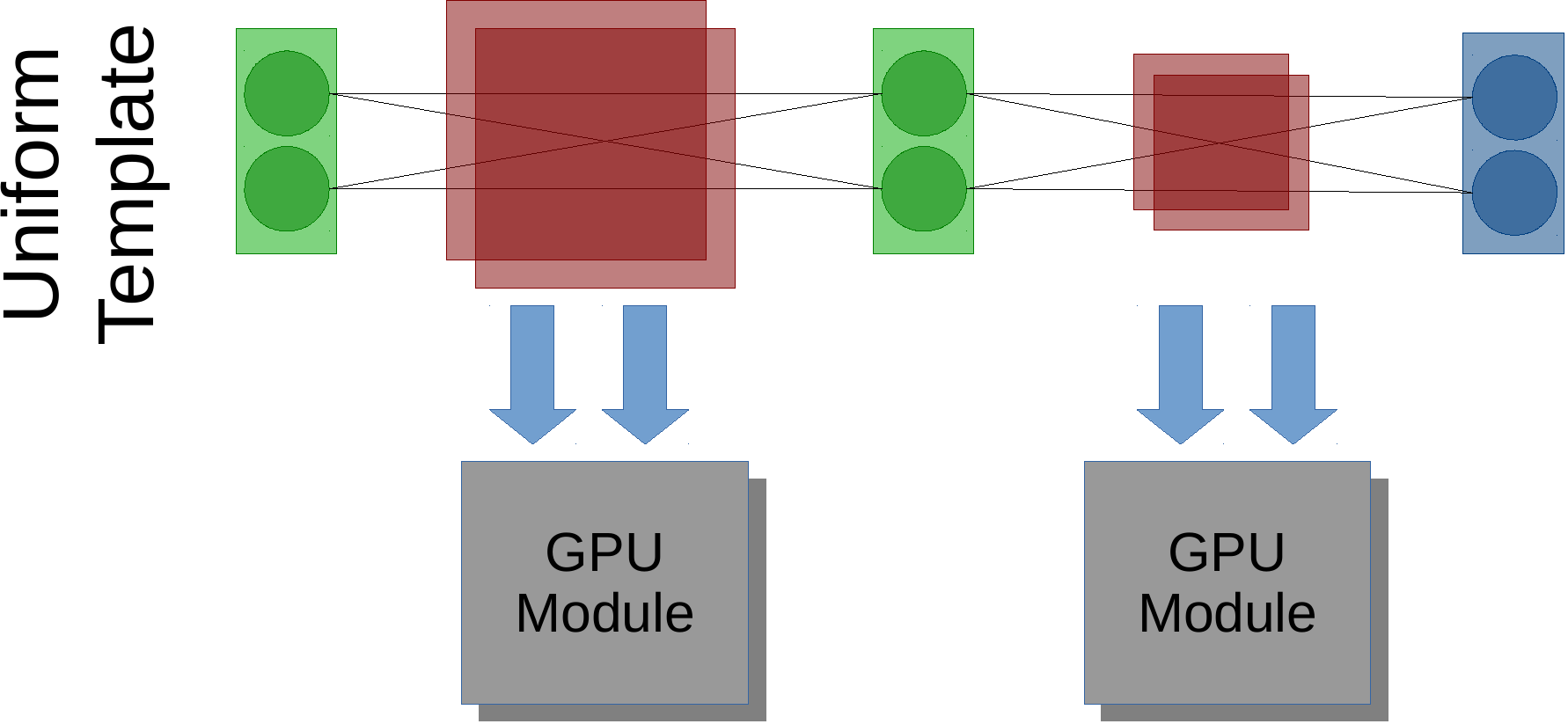}\par
  \end{multicols}
    \caption{A toy example to show how two different templates with the same number of filters produce a variety of effects in parameters, memory, inference time and flops. Layers (rectangles) contain in total, equal number of filters (circles) for both templates. Lines between filters represent parameters, red squares are by-channel feature maps which reside in memory jointly with parameters. Flops are produced by shifting filters along feature maps. Inference time is affected by flops and number of transfers, indicated by blue arrows and here limited to two simultaneously, between memory and GPU modules. Diagram assumes filters of equal sizes and pooling between layers. Differences are scaled up in real models counting thousand of filters.}
\label{fig:templates_resource_change}
\end{figure}

Different distributions with the same number of filters can lead to different number of parameters (e.g. weights) and different memory or computational requirements (e.g. GPU modules). In the toy example in Figure \ref{fig:templates_resource_change}, both models have the same number of filters but the one on the right has less parameters and less compute requirements at the cost of more memory footprint. This example highlights the compromises that filter distributions can offer for the design and operation of network models.

We define a convolutional neural network base model as a set of numbered layers $\textsl{L}=1,... ,D+1$, each with $f_{l}$ filters in layer $l$, $D+1$ is the final classification layer. The total number of filters that can be redistributed is given by $F = \sum_{l=1}^{D}f_{l}$. We want to test if the common heuristic of distributing $F$ having $f_{l+1} = 2f_{l}$ each time the feature map is halved, is advantageous to the model over other distributions of $F$ when evaluating performance, memory footprint and inference time.

The number of filters in the final layer $D+1$ depends on the task and remains the same for all the templates, therefore it is not taken into account for computing the number of filters to redistribute. For architectures composed of blocks (e.g. Inception) we consider blocks as single layers and keep the number of filters within a block the same. As a result, a final Inception module marked with $f_{l}$ filters, is set to that number of filters in each layer inside the module.

\textbf{Uniform Template}. The most simple distribution to evaluate is, as the original Neocognitron, an uniform distribution of filters. Computing the number of filters in an uniform distribution is straightforward, the new number in each layer is given by $f'_{l}= F / D  \quad  \forall l \in \left \{1,... ,D  \right \}$.

\textbf{Reverse Template}. Another straight-forward transformation for the filter distribution adopted in this paper is reversing the number of filters in every layer. Our final model with this template is defined by the filters $f'_{l} = f_{D-l+1}$.

\textbf{Quadratic Template}. This distribution is characterised by a quadratic equation $f'_{l} = al^{2}+bl + c$ and consequently, has a parabolic shape with the vertex in the middle layer. We set this layer to the minimal number of filters in the base model $f_{min} = min\left ( f_{l}  \right )  \quad  l \in \left \{1,... ,D  \right \}$ so, the number of filters is described by  $f'_{D/2} = f_{min}$. Also, we find the maximum value in both the initial and final convolutional layers, thus $f'_{1} = f'_{D}$.

To compute the new number of filters in each layer we solve the system of linear equations given by \textit{i)} the restriction of the total number of filters in $\sum_{l=1}^{D}\left ( f'_{l} \right ) = \sum_{l=1}^{D}\left ( al^{2}+bl + c \right ) = F$, that can be reduced to $\left ( \frac{D^{3}}{3} + \frac{D^{2}}{2} + \frac{D}{6} \right )a + \left ( \frac{D^{2}}{2} + \frac{D}{2} \right )b + Dc = F$, \textit{ii)} the equation produced by the value in the vertex $f'_{D/2} = \frac{D}{2}^{2}a+\frac{D}{2}b + c = f_{min}$ and \textit{iii)} the equality from the maximum values which reduces to $(D^2-1)a +(D-1)b = 0$.

\textbf{Negative Quadratic Template}. It is a parabola with the vertex in a maximum, that is, a negative quadratic curve. The equation is the same as the previous template but the restrictions change. Instead of defining a value in the vertex, $f'_{l}$ at the initial and final convolutional layers are set to the minimal number of filters in the base model $f'_{l} = f_{min} \quad l \in \left \{1,D   \right \}$. 
The number of filters in each layer is computed again with a system of equations specified by \textit{i)}, the restriction of the total number of filters as in the quadratic template, and the two points already known in the first and last convolutional layers defined by \textit{ii)} $a + b + c = f_{min}$ and \textit{iii)} $D^{2}a + Db + c = f_{min}$.

%%%%%%%%%%%%%%%%%%%%%%%%%%%%%%%%%%%%%%%%%%%%%%%%%%%%%%%%%%%%%%%%%%%%%%%%%%%%%%%%
\section{Models Comparison Under Size, Memory Footprint and Speed}\label{chap:experiments}

In this section we investigate the effects of applying different templates to the distribution of kernels in convolutional neural network models (VGG, ResNet, Inception and MobileNet). We compare models under the basis of size, memory and speed in three of the popular datasets for classification tasks.\\

\begin{table}
\caption{Model performances with the original distribution and four templates for the same number of filters evaluated on CIFAR-10, CIFAR-100 and Tiny-Imagenet datasets. After filter redistribution models surpass the base accuracy. Results show average of three repetitions.}
\footnotesize
\begin{tabularx}{\linewidth}{|c|C|C|C|C|C|}
\hline
           &                & \multicolumn{4}{c|}{Redistribution Templates}                         \\ \cline{3-6} 
Model     & Base           & Rev Base   & Unif        & Quad      & Neg Quad \\ \hline
\multicolumn{6}{|l|}{CIFAR-10} \\ \hline 
 VGG-19    & 93.52          & \textbf{94.40} & 94.24          & 94.18          & 94.21              \\ \cline{1-6} 
 ResNet-50 & 94.70          & 95.17          & 95.08          & 94.41          & \textbf{95.23}     \\ \cline{1-6} 
 Inception & 94.84          & 94.60          & 94.82          & \textbf{94.86} & 94.77              \\ \cline{1-6} 
 MobileNet & 89.52          & \textbf{91.35} & 91.28          & 89.98          & 91.04              \\ \hline
 \multicolumn{6}{|l|}{CIFAR-100} \\ \hline 
 VGG-19    & 71.92          & \textbf{74.65} & 74.03          & 73.55          & 74.05              \\ \cline{1-6} 
 ResNet-50 & \textbf{77.09} & 74.80          & 76.65          & 75.71          & 76.76              \\ \cline{1-6} 
 Inception & 78.03          & 77.78          & \textbf{78.12} & 77.67          & 76.65              \\ \cline{1-6} 
 MobileNet & 65.08          & 66.39          & \textbf{68.71} & 63.89          & 67.05              \\ \hline
 \multicolumn{6}{|l|}{Tiny-Imagenet} \\ \hline 
 VGG-19    & 54.62          & 57.73          & 56.68          & 54.73          & \textbf{59.50}     \\ \cline{1-6} 
 ResNet-50 & \textbf{61.52} & 53.67          & 60.97          & 59.77          & 60.12              \\ \cline{1-6} 
 Inception & 54.80          & 55.24          & 55.78          & 54.97          & \textbf{55.87}     \\ \cline{1-6} 
 MobileNet & 56.29          & 51.40          & \textbf{58.11} & 53.37          & 55.76              \\ \hline
\end{tabularx}
\label{tab:templates_accuracy}
\end{table}

\begin{table*}[hbt!]
\caption{Parameters, memory and inference time for selected models when applying our templates keeping the same number of filters evaluated on the CIFAR-10 (black) and Tiny-Imagenet (blue) datasets. Models are normally optimised to fast GPU operation, therefore the original base distribution has a good effect in speed but the redistribution of filters induced by our templates makes models capabilities improve on the other metrics. Memory footprint reported by CUDA.}
\footnotesize
\begin{tabularx}{\textwidth}{|l|C|r|>{\color{blue}}r|r|>{\color{blue}}r|r|>{\color{blue}}r|r|>{\color{blue}}r|r|>{\color{blue}}r|}   
\hline
 & & \multicolumn{2}{C|}{}     & \multicolumn{8}{c|}{Redistribution Templates}                                                     \\  \cline{5-12} 
  Resource              & Model     & \multicolumn{2}{C|}{Base} & \multicolumn{2}{C|}{Reverse Base} & \multicolumn{2}{C|}{Uniform} & \multicolumn{2}{C|}{Quadratic} & \multicolumn{2}{C|}{Negative Quadratic} \\ \hline 

                  & VGG-19            & 20.0          & 25.0          & 20.0            & 20.6            & 16.0          & \textbf{19.3} & \textbf{15.8}  & 20.7          & 20.0                & 20.6              \\ \cline{2-12} 
Parameters        & ResNet-50         & 23.5          & 23.9          & 23.1            & 23.1            & \textbf{12.9} & \textbf{13.0} & 19.0           & 19.3          & 33.0                & 33.0              \\ \cline{2-12} 
(Millions)        & Inception         & \textbf{6.2}  & 19.2          & 6.7             & \textbf{10.0}   & \textbf{6.2}  & 12.7          & 7.2            & 18.7          & 7.0                 & 10.1              \\ \cline{2-12} 
                  & MobileNet         & 3.2           & 3.4           & \textbf{2.2}    & \textbf{2.4}    & \textbf{2.2}  & \textbf{2.4}  & 3.2            & 3.3           & 2.4                 & 2.6               \\ \hline
Memory            & VGG-19            & \textbf{1.3}  & 1.5           & 2.6             & 10.0            & 4.4           & 4.8           & 2.0            & 6.8           & 1.4                 & 3.8               \\ \cline{2-12} 
Footprint         & ResNet-50         & 3.1           & 5.0           & 11.5            & 10.1            & 4.1           & 9.6           & 7.9            & 7.5           & \textbf{3.0}        & 9.8               \\ \cline{2-12} 
(GB/batch)        & Inception         & \textbf{1.5}  & 5.8           & 3.1             & 10.8            & 1.7           & 6.7           & 2.2            & 8.6           & 1.6                 & 5.9               \\ \cline{2-12} 
                  & MobileNet         & 2.5           & 2.5           & 5.1             & 5.1             & 1.5           & 5.9           & 6.0            & 4.8           & \textbf{1.0}        & \textbf{1.9}      \\ \hline
Inference         & VGG-19            & \textbf{3.0}  & 4.9           & 8.2             & 4.1             & 5.3           & 4.2           & 7.5            & 4.6           & 7.3                 & \textbf{3.5}      \\ \cline{2-12} 
Time              & ResNet-50         & 46.4          & 13.3          & 61.0            & 12.8            & \textbf{23.4} & 12.8          & 59.0           & \textbf{11.0} & 47.6                & 29.9              \\ \cline{2-12} 
(ms/batch)        & Inception         & 28.5          & 24.0          & 54.9            & 21.4            & 34.3          & 28.3          & 25.2           & \textbf{18.3} & \textbf{24.3}       & 31.4              \\ \cline{2-12} 
                  & MobileNet         & \textbf{3.8}  & 5.8           & 6.8             & 6.7             & 4.3           & 9.7           & 7.4            & 7.3           & 4.9                 & \textbf{5.3}      \\ \hline
\end{tabularx}

\label{tab:templates_parameters}
\end{table*}

%%%%%%%%%%%%%%%%%%%%%%%%%%%%%%%%%%%%%%%%%%%%%%%%%%%%%%%%%%%%%%%%%%%%%%%%%%%%%%%%

\subsection*{Datasets and Models}

We trained over three datasets traditionally used for convolutional network evaluation: CIFAR-10, CIFAR-100 \cite{krizhevsky2009learning} and Tiny-Imagenet \cite{le2015tiny}. The first two datasets contain sets of 50,000 and 10,000 colour images for train and validation respectively, with a resolution of 32x32. Tiny-Imagenet is a reduced version of the original Imagenet dataset with only 200 classes and images with a resolution of 64 x 64 pixels.

We evaluate some of the most popular CNN models: VGG\cite{simonyan2014very}, ResNet\cite{he2016deep}, Inception\cite{szegedy2015going} and MobileNet\cite{howard2017mobilenets}; which represent some of the highest performing CNNs on the ImageNet challenge in previous years \cite{russakovsky2015imagenet}.

\subsection*{Implementation Details}

Experiments have models fed with images with the common augmentation techniques of padding, random cropping and horizontal flipping. Our experiments were run in a NVidia Titan X Pascal 12GB GPU adjusting the batch size to 128.
All convolutional models, with and without templates, were trained for 160 epochs using the same conditions. Therefore, there is some margin for improving accuracy for each distribution by performing individual hyperparameter \cite{mittal2020hyperstar,li2017hyperband}. We used stochastic gradient descent (SGD) with weight decay of 1e-4, momentum of 0.9 and a scheduled learning rate starting in 0.1 for the first 80 epochs, 0.01 for the next 40 epochs and finally 0.001 for the remaining epochs.

\subsection*{Template effect over baseline models}

We conducted an experiment to test our proposed templates on the selected architectures. Table \ref{tab:templates_accuracy} shows VGG, Inception and MobileNet accuracies improving in all datasets when templates are applied. Being complex architectures, ResNet and Inception present the highest accuracy in general. A surprising finding is that in both models difference in accuracy between templates is less than 2.3\% despite the drastic modifications that models are suffering after the change of filter distribution. Models that share a sequential classical architecture such as VGG and MobileNet, show a better improvement when using templates in Tiny-Imagenet. A remarkable accuracy improvement of 4.88 percentage point is achieved in VGG.

When analysing resource consumption (Table \ref{tab:templates_parameters}), we find models are affected differently with each template and model. Reverse-Base, Uniform and Quadratic templates show some reductions in the number of parameters while Negative Quadratic template reduces the memory usage. Inference Time is affected negatively for most of the templates. This is an expected result as original models are designed to perform well in the GPU. Inception model shows an improvement in speed with reductions of 14\% over inference time respect to the base model while maintaining comparable accuracy. ResNet is able to reduce inference time in 49\% at the cost of having slightly less accuracy than the base model.

%%%%%%%%%%%%%%%%%%%%%%%%%%%%%%%%%%%%%%%%%%%%%%%%%%%%%%%%%%%%%%%%%%%%%%%%%%%%%%%%
\section{Conclusions}\label{chap:conclusions} 

The most common design of convolutional neural networks when choosing the distribution of the number of filters is to start with a few and then to increase the number in deeper layers. We challenged this design by evaluating some architectures with a varied set of distributions on the CIFAR and Tiny-Imagenet datasets. Our results suggest that this pyramidal distribution is not necessarily the best option for obtaining the highest accuracy or even the highest parameter efficiency.

Our experiments show that models, with the same amount of filters but different distributions produced by our templates, improve accuracy with up to 4.8 points for some model-task pairs. In terms of resource consumption, they can obtain a competitive accuracy compared to the original models using less resources with up to 56\% less parameters and a memory footprint up to 60\% smaller. Results also reveal an interesting behaviour in evaluated models: a strong resilience to changes in filter distribution. The variation in accuracy for all models after administering templates is less than 5\% despite the considerable modifications in the distributions and therefore, in the original design. Our work overall offers insights to model designers, both automated and manual, to construct more efficient models by introducing the idea of new distributions of filters for neural network models and help gather data to build understanding of the design process for model-task pairs.\\

\subsubsection*{Acknowledgments}
\AcknowledgmentsText

\clearpage
\clearpage

\bibliographystyle{ieee_fullname}
\bibliography{bibliography}

\begin{thebibliography}{10}\itemsep=-1pt

\bibitem{chu2014analysis}
Joseph~Lin Chu and Adam Krzy{\.z}ak.
\newblock Analysis of feature maps selection in supervised learning using
  convolutional neural networks.
\newblock In {\em Canadian Conference on Artificial Intelligence}, pages
  59--70. Springer, 2014.

\bibitem{dong2019network}
Xuanyi Dong and Yi Yang.
\newblock Network pruning via transformable architecture search.
\newblock {\em arXiv preprint arXiv:1905.09717}, 2019.

\bibitem{frankle2018lottery}
Jonathan Frankle and Michael Carbin.
\newblock The lottery ticket hypothesis: Finding sparse, trainable neural
  networks.
\newblock {\em arXiv preprint arXiv:1803.03635}, 2018.

\bibitem{fukushima1980neocognitron}
Kunihiko Fukushima.
\newblock Neocognitron: A self-organizing neural network model for a mechanism
  of pattern recognition unaffected by shift in position.
\newblock {\em Biological cybernetics}, 36(4):193--202, 1980.

\bibitem{gordon2018morphnet}
Ariel Gordon, Elad Eban, Ofir Nachum, Bo Chen, Hao Wu, Tien-Ju Yang, and Edward
  Choi.
\newblock Morphnet: Fast \& simple resource-constrained structure learning of
  deep networks.
\newblock In {\em Proceedings of the IEEE Conference on Computer Vision and
  Pattern Recognition}, pages 1586--1595, 2018.

\bibitem{he2016deep}
Kaiming He, Xiangyu Zhang, Shaoqing Ren, and Jian Sun.
\newblock Deep residual learning for image recognition.
\newblock In {\em Proceedings of the IEEE conference on computer vision and
  pattern recognition}, pages 770--778, 2016.

\bibitem{he2019filter}
Yang He, Ping Liu, Ziwei Wang, Zhilan Hu, and Yi Yang.
\newblock Filter pruning via geometric median for deep convolutional neural
  networks acceleration.
\newblock In {\em Proceedings of the IEEE Conference on Computer Vision and
  Pattern Recognition}, pages 4340--4349, 2019.

\bibitem{howard2017mobilenets}
Andrew~G Howard, Menglong Zhu, Bo Chen, Dmitry Kalenichenko, Weijun Wang,
  Tobias Weyand, Marco Andreetto, and Hartwig Adam.
\newblock Mobilenets: Efficient convolutional neural networks for mobile vision
  applications.
\newblock {\em arXiv preprint arXiv:1704.04861}, 2017.

\bibitem{izquierdo2021filter}
Ramon Izquierdo-Cordova and Walterio Mayol-Cuevas.
\newblock Towards efficient convolutional network models with filter
  distribution templates.
\newblock {\em arXiv preprint arXiv:2104.08446}, 2021.

\bibitem{krizhevsky2009learning}
Alex Krizhevsky, Geoffrey Hinton, et~al.
\newblock Learning multiple layers of features from tiny images.
\newblock Technical report, Citeseer, 2009.

\bibitem{le2015tiny}
Ya Le and Xuan Yang.
\newblock Tiny imagenet visual recognition challenge.
\newblock {\em CS 231N}, 7:7, 2015.

\bibitem{leclerc2018smallify}
Guillaume Leclerc, Manasi Vartak, Raul~Castro Fernandez, Tim Kraska, and Samuel
  Madden.
\newblock Smallify: Learning network size while training.
\newblock {\em arXiv preprint arXiv:1806.03723}, 2018.

\bibitem{lecun1998gradient}
Yann LeCun, L{\'e}on Bottou, Yoshua Bengio, Patrick Haffner, et~al.
\newblock Gradient-based learning applied to document recognition.
\newblock {\em Proceedings of the IEEE}, 86(11):2278--2324, 1998.

\bibitem{li2017hyperband}
Lisha Li, Kevin Jamieson, Giulia DeSalvo, Afshin Rostamizadeh, and Ameet
  Talwalkar.
\newblock Hyperband: A novel bandit-based approach to hyperparameter
  optimization.
\newblock {\em The Journal of Machine Learning Research}, 18(1):6765--6816,
  2017.

\bibitem{li2019learning}
Yawei Li, Shuhang Gu, Luc~Van Gool, and Radu Timofte.
\newblock Learning filter basis for convolutional neural network compression.
\newblock In {\em Proceedings of the IEEE International Conference on Computer
  Vision}, pages 5623--5632, 2019.

\bibitem{liu2018rethinking}
Zhuang Liu, Mingjie Sun, Tinghui Zhou, Gao Huang, and Trevor Darrell.
\newblock Rethinking the value of network pruning.
\newblock {\em arXiv preprint arXiv:1810.05270}, 2018.

\bibitem{mittal2020hyperstar}
Gaurav Mittal, Chang Liu, Nikolaos Karianakis, Victor Fragoso, Mei Chen, and
  Yun Fu.
\newblock Hyperstar: Task-aware hyperparameters for deep networks.
\newblock In {\em Proceedings of the IEEE/CVF Conference on Computer Vision and
  Pattern Recognition}, pages 8736--8745, 2020.

\bibitem{ren2020comprehensive}
Pengzhen Ren, Yun Xiao, Xiaojun Chang, Po-Yao Huang, Zhihui Li, Xiaojiang Chen,
  and Xin Wang.
\newblock A comprehensive survey of neural architecture search: Challenges and
  solutions.
\newblock {\em arXiv preprint arXiv:2006.02903}, 2020.

\bibitem{russakovsky2015imagenet}
Olga Russakovsky, Jia Deng, Hao Su, Jonathan Krause, Sanjeev Satheesh, Sean Ma,
  Zhiheng Huang, Andrej Karpathy, Aditya Khosla, Michael Bernstein, et~al.
\newblock Imagenet large scale visual recognition challenge.
\newblock {\em International Journal of Computer Vision}, 115(3):211--252,
  2015.

\bibitem{simonyan2014very}
Karen Simonyan and Andrew Zisserman.
\newblock Very deep convolutional networks for large-scale image recognition.
\newblock {\em arXiv preprint arXiv:1409.1556}, 2014.

\bibitem{szegedy2015going}
Christian Szegedy, Wei Liu, Yangqing Jia, Pierre Sermanet, Scott Reed, Dragomir
  Anguelov, Dumitru Erhan, Vincent Vanhoucke, and Andrew Rabinovich.
\newblock Going deeper with convolutions.
\newblock In {\em Proceedings of the IEEE conference on computer vision and
  pattern recognition}, pages 1--9, 2015.

\bibitem{wang2020revisiting}
Jiaxing Wang, Haoli Bai, Jiaxiang Wu, Xupeng Shi, Junzhou Huang, Irwin King,
  Michael Lyu, and Jian Cheng.
\newblock Revisiting parameter sharing for automatic neural channel number
  search.
\newblock {\em Advances in Neural Information Processing Systems}, 33, 2020.

\bibitem{you2019gate}
Zhonghui You, Kun Yan, Jinmian Ye, Meng Ma, and Ping Wang.
\newblock Gate decorator: Global filter pruning method for accelerating deep
  convolutional neural networks.
\newblock In {\em Advances in Neural Information Processing Systems}, pages
  2130--2141, 2019.

\bibitem{zoph2017learning}
Barret Zoph, Vijay Vasudevan, Jonathon Shlens, and Quoc~V Le.
\newblock Learning transferable architectures for scalable image recognition.
\newblock {\em arXiv preprint arXiv:1707.07012}, 2017.

\bibitem{zoph2018learning}
Barret Zoph, Vijay Vasudevan, Jonathon Shlens, and Quoc~V Le.
\newblock Learning transferable architectures for scalable image recognition.
\newblock In {\em Proceedings of the IEEE conference on computer vision and
  pattern recognition}, pages 8697--8710, 2018.

\end{thebibliography}

\end{document}